\pdfoutput=1

\documentclass[11pt]{article}

\usepackage{acl}

\usepackage{times}
\usepackage{latexsym}
\usepackage{graphicx}
\usepackage{soul}
\usepackage{amsmath}
\usepackage{color}
\usepackage{xspace}
\usepackage{bm}
\usepackage{amsfonts}
\usepackage[T1]{fontenc}
\usepackage{adjustbox}
\usepackage{booktabs}
\usepackage{pbox}
\usepackage{multirow, multicol}
\usepackage[colorinlistoftodos,prependcaption]{todonotes}
\usepackage{tabulary,booktabs}
\usepackage{todonotes}
\usepackage{subfig}
\usepackage{tabulary,booktabs}
\usepackage{tablefootnote}
\usepackage{amsfonts}
\usepackage{multirow}
\usepackage{colortbl}
\usepackage{color, xcolor}
\usepackage[subtle]{savetrees}
\usepackage{comment}
\usepackage{amsmath}
\usepackage{color,soul}
\usepackage{float}
\usepackage{pifont}
\usepackage{chngpage}
\usepackage{mathtools}
\usepackage{epigraph}
\usepackage{multirow}
\usepackage{xcolor,colortbl}
\usepackage{bm}
\usepackage{makecell}
\usepackage{multirow}

\usepackage[utf8]{inputenc}

\usepackage{microtype}
\usepackage{float}
\usepackage{xcolor}
\usepackage{booktabs}
\usepackage{graphicx}
\usepackage{tcolorbox}
\usepackage{multirow}

\usepackage[colorinlistoftodos,prependcaption]{todonotes}
\newcommand{\todocomment}[1]{} 

\newcommand{\jh}[1]{\textcolor{olive}{\todocomment{~jh: #1}}}
\newcommand{\rk}[1]{} 
\newcommand{\change}[1]{\textcolor{black}{#1}}

\newcommand{\remove}[1]{}
\newcommand{\textoverset}[2]{\ensuremath{\mathop{\kern\z\mbox{#2}}\limits^{\mbox{#1}}}}
\newcommand{\perfect}{\textsc{Perfect}\xspace}
\newcommand{\finetune}{\textsc{Finetune}\xspace}

\newcommand{\pet}{\textsc{PET}\xspace}

\newcommand{\argmax}{\operatornamewithlimits{argmax}}
\newcommand{\mask}{\textsc{[MASK]}\xspace}
\newcommand{\smaller}[1]{{\scriptsize #1}}

\newcommand{\drawbox}[1]{ {\setlength{\fboxrule}{0.8pt}\fcolorbox{black}{white!95!black}{\makebox[5cm][l]{#1}}}}
\definecolor{Gray}{gray}{0.85}
\definecolor{LightCyan}{rgb}{0.88,1,1}
\newcolumntype{a}{>{\columncolor{Gray}}c}
\newcolumntype{b}{>{\columncolor{white}}c}
\newcolumntype{H}{>{\setbox0=\hbox\bgroup}c<{\egroup}@{}}

\title{
\perfect: 
Prompt-free and Efficient Few-shot Learning with Language Models
}

\newcommand{\affilsup}[1]{\rlap{\textsuperscript{\normalfont#1}}}
\author{
    Rabeeh Karimi Mahabadi\affilsup{1,3,4} \qquad
    Luke Zettlemoyer\affilsup{1,2} \qquad
    James Henderson\affilsup{4}\\%
    \textbf{Marzieh Saeidi}\affilsup{1} \qquad
    \textbf{Lambert Mathias}\affilsup{1} \qquad
    \textbf{Veselin Stoyanov}\affilsup{1} \qquad
    \textbf{Majid Yazdani}\affilsup{1} \\ 
    $^1$Meta AI \qquad
    $^2$University of Washington \qquad
    $^3$EPFL \qquad 
    $^4$Idiap Research Institute \\
    \texttt{\{rabeeh.karimi, james.henderson\}@idiap.ch}\\
    \texttt{lsz@cs.washington.edu} \\
    \texttt{\{marzieh,mathiasl,ves,myazdani\}@fb.com} 
}

\begin{document}
\maketitle

\begin{abstract}
Current methods for few-shot fine-tuning of pretrained masked language models (PLMs) require carefully engineered prompts and verbalizers for each new task to convert examples into a cloze-format that the PLM can score. In this work, we propose \perfect, a simple and efficient method for few-shot fine-tuning of PLMs \emph{without relying on any such handcrafting}, which is highly effective given as few as 32 data points. \perfect makes two key design choices: First, we show that manually engineered task prompts can be replaced with \emph{task-specific adapters} that enable sample-efficient fine-tuning and reduce memory and storage costs by roughly factors of 5 and 100, respectively. Second, instead of using handcrafted verbalizers, we learn new \emph{multi-token label embeddings} during fine-tuning, which are not tied to the model vocabulary and which allow us to avoid complex auto-regressive decoding. These embeddings are not only learnable from limited data but also enable nearly 100x faster training and inference. 
Experiments on a wide range of few shot NLP tasks demonstrate that \perfect, while being simple and efficient, also outperforms existing state-of-the-art few-shot learning methods. Our code is publicly available at \url{https://github.com/facebookresearch/perfect.git}.
\end{abstract}

\section{Introduction} 
\begin{figure}[tp]
\centering 
\includegraphics[width=0.45\textwidth]{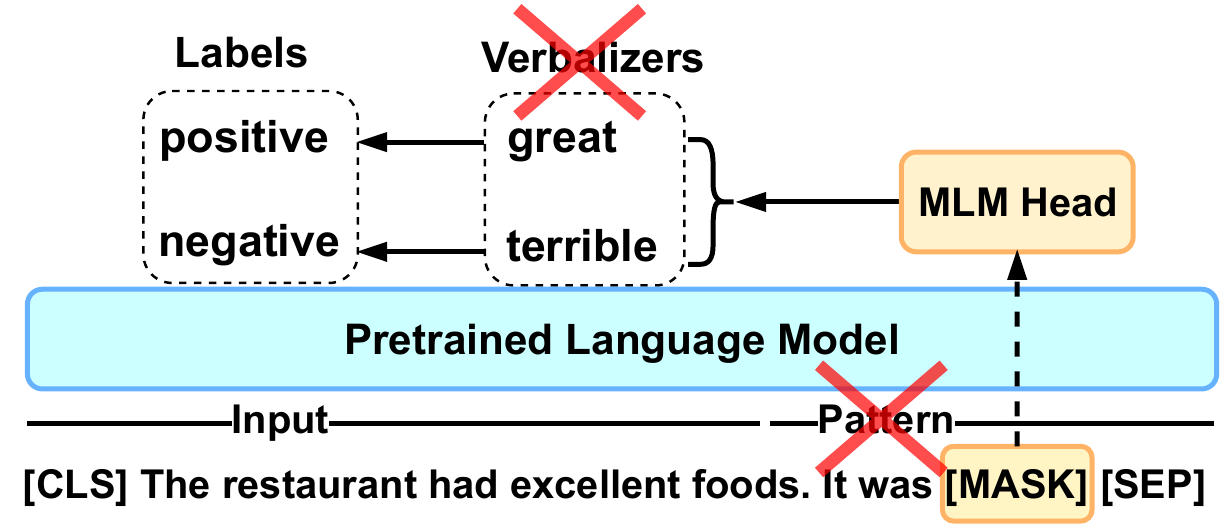}
\caption{\change{Existing few-shot fine-tuning methods require manual engineering to reduce new tasks to masked language modeling. \perfect does not rely on any handcrafting, removing both patterns and verbalizers (see Figure~\ref{fig:perfect}).}}
\label{fig:pet} 
\end{figure}

Recent methods for few-shot language model tuning obtain impressive performance but require careful engineering of prompts and verbalizers to convert inputs to a cloze-format~\citep{taylor1953cloze} that can be scored with pre-trained language models (PLMs)~\citep{radford2018improving, radfordlanguage, brown2020language, PET1, PET2}.  For example, as Figure~\ref{fig:pet} shows, a sentiment classifier can be designed by inserting the input text $\bm{x}$ in a {\em prompt template} ``$\bm{x}$ It was \mask'' where  {\em verbalizers} (e.g., `great' and `terrible') are substituted for the \mask to score target task labels (`positive' or `negative'). In this paper, we show that such engineering is not needed for few-shot learning and instead can be replaced with simple methods for data-efficient fine-tuning with as few as 32 end-task examples. 

More specifically, we propose \perfect, a 
Prompt-free and Efficient paRadigm for FEw-shot Cloze-based fine-Tuning.
To remove handcrafted patterns, \perfect uses \emph{task-specific adapter layers} \cite{houlsby2019parameter, pfeiffer2020adapterhub} (\textsection\ref{sec:pattern_free}). Freezing the underlying PLM with millions or billions of parameters~\citep{liu2019roberta, raffel2020exploring}, and only tuning adapters with very few new parameters saves on memory and storage costs (\textsection\ref{sec:efficiency}), while allowing very sample-efficient tuning (\textsection\ref{sec:experiments}). It also  stabilizes the training by increasing the worst-case performance and decreasing variance across the choice of examples in the few shot training sets (\textsection\ref{sec:analysis}). 

To remove handcrafted verbalizers (with variable token lengths), we introduce a new \emph{multi-token fixed-length classifier scheme} that learns task label embeddings which are independent from the language model vocabulary during fine-tuning (\textsection\ref{sec:soft_verbalizers}). We show (\textsection\ref{sec:experiments}) that this approach is sample efficient and outperforms carefully engineered verbalizers from \emph{random initialization} (\textsection\ref{sec:experiments}).
It also allows us to avoid previously used expensive auto-regressive decoding schemes~\citep{PET2}, by leveraging prototypical networks~\citep{snell2017prototypical} over multiple tokens. 
Overall, these changes enable up to 100x faster learning and inference (\textsection\ref{sec:efficiency}).

\perfect has several advantages: It avoids engineering patterns and verbalizers for each new task, which can be cumbersome. Recent work has shown that even some intentionally irrelevant or misleading prompts can perform as well as more interpretable ones~\cite{webson2021prompt}. 
Unlike the zero-shot or extreme few-shot case, where prompting might be essential, we argue in this paper that all you need is tens of training examples to avoid these challenges by adopting \perfect or a similar data-efficient learning method. Experiments on a wide variety of NLP tasks demonstrate that \perfect outperforms state-of-the-art prompt-based methods while being significantly more efficient in inference and training time, storage, and memory usage (\textsection\ref{sec:efficiency}). To the best of our knowledge, we are the first to propose a few-shot learning method using the MLM objective in PLMs that provide state-of-the-art results while removing all per-task manual engineering.

\section{Background}
\paragraph{Problem formulation:} We consider a general problem of fine-tuning language models in a few-shot setting, on a small training set with $K$ unique classes and $N$ examples per class, such that the total number of examples is $|\mathcal{D}|=N \times K$. Let 
$\mathcal{D} = \cup_{k=1}^K \mathcal{D}_k$  
be the given training set, where  $\mathcal{D}_k= \{(\bm{x^i_k}, y^i_k)\}_{i=1}^N$ shows the set of examples labeled with class $k$ and $y^i_k \in \mathcal{Y}$ is the corresponding label, where $|\mathcal{Y}|=K$. We additionally assume access to a development set with the same size as the training data. Note that larger validation sets can grant a substantial advantage \citep{perez2021true}, and thus it is important to use a limited validation size to be in line with the goal of few-shot learning. Unless specified otherwise, in this work, we use 16 training examples ($N = 16$) and a validation set with 16 examples, for a total of 32-shot learning.

\subsection{Adapters} \label{sec:adapters}
\begin{figure}[tp]
\centering 
\includegraphics[width=0.4\textwidth]{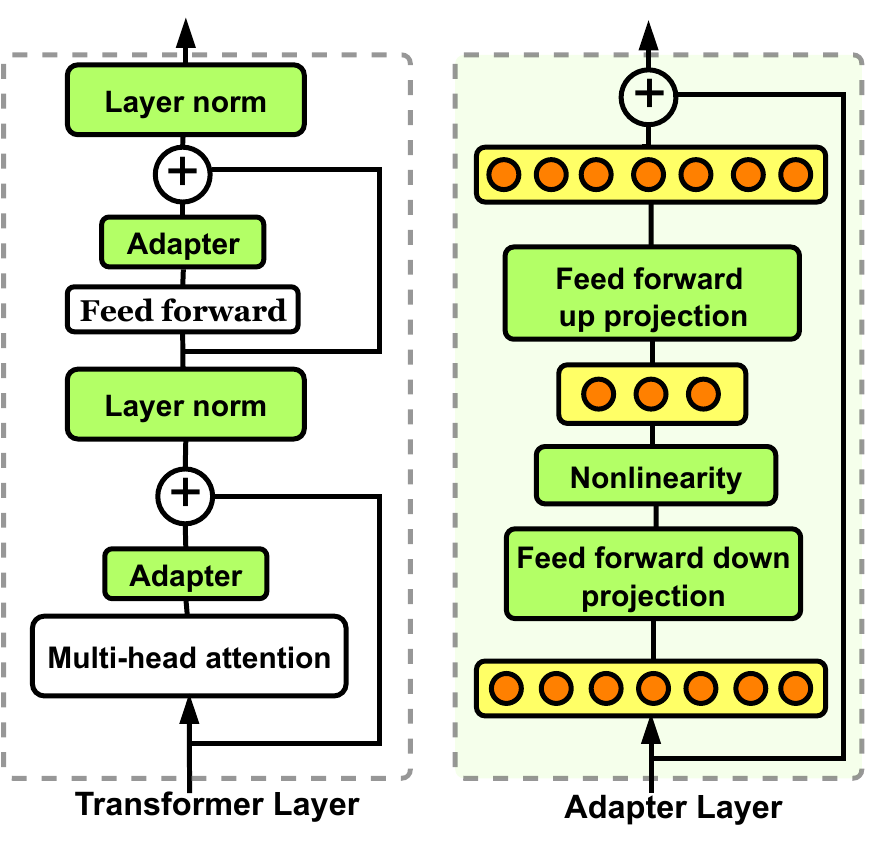}
\caption{
Left: Adapter integration in a PLM.  Right: An adapter architecture. Adapters are usually inserted after the feed-forward and self-attention modules. During training, we only optimize the green components}
\label{fig:our_method} 
\end{figure}

Recent work has shown that fine-tuning \emph{all} parameters of PLMs with a large number of parameters in low-resource datasets can lead to a sub-optimal solution~\citep{peters-2019-tune, dodge2020fine}. As shown in Figure~\ref{fig:our_method}, \citet{rebuffi2018efficient} and \citet{houlsby2019parameter} suggest an efficient alternative, by inserting small task-specific modules called \emph{adapters} within layers of a PLMs. They then only train the newly added adapters and layer normalization, while fixing the remaining parameters of a PLM.

Each layer of a transformer model is composed of two primary modules: a) an attention block, and b) a feed-forward block, where both modules are followed by a skip connection.  As depicted in Figure~\ref{fig:our_method}, adapters are normally inserted after each of these blocks before the skip connection.  

Adapters are bottleneck architectures. By keeping input and output dimensions the same, they introduce no additional architectural changes. Each adapter, $A(.) \in \mathbb{R}^H$, consists of a down-projection, $D(.)\in\mathbb{R}^{H\times B}$, a non-linearity, such as GeLU~\citep{hendrycks2016gaussian}, and an up-projection $U(.)\in\mathbb{R}^{B\times H}$, where $H$ is the dimension of input hidden states $\bm{x}$, and $B$ is the bottleneck size. Formally defined as:
\begin{align}
A(\bm{x})=U(\text{GeLU}(D(\bm{x}))) + \bm{x}, \label{eq:adapters}
\\[-4ex]\nonumber
\end{align}

\subsection{Prompt-based Fine-tuning}
\paragraph{Standard Fine-tuning:} In standard fine-tuning with PLMs~\citep{devlin-etal-2019-bert}, first a special $\text{[CLS]}$ token is appended to the input $\bm{x}$, and then the PLM maps it to a sequence of hidden representations $\bm{h}=(\bm{h_1},\dots,\bm{h_S})$ with $\bm{h_i}\in \mathbb{R}^{H}$,  where $H$ is the hidden dimension, and $S$ is the maximum sequence length.
Then, a classifier, $\textit{softmax}(\bm{W^T} \bm{h}_{\text{[CLS]}})$, using the embedding of the classification token ($\bm{h}_{\text{[CLS]}}$), is trained end-to-end for each downstream task. The main drawback of this approach is the discrepancy between the pre-training and fine-tuning phases since PLMs have been trained to \emph{predict mask tokens} in a masked language modeling task~\citep{devlin-etal-2019-bert}.

\paragraph{Prompt-based tuning:} To address this discrepancy, \emph{prompt-based fine-tuning}~\citep{PET1, PET2, gao2020making} formulates tasks in a cloze-format~\citep{taylor1953cloze}. This way, the model can predict targets with a \emph{masked language modeling (MLM) objective}. For example, as shown in Figure~\ref{fig:pet}, for a sentiment classification task, inputs are converted to:
\begin{align}
\bm{x}_\text{prompt} ~=~ \text{[CLS]}~ \bm{x}~ .~  \underbrace{\text{It was}}_{\text{pattern}}~ \text{[MASK]}~.~\text{[SEP]} \nonumber 
\end{align}
Then, the PLM determines which {\em verbalizer} (e.g., `great' and `terrible') is the most likely substitute for the mask in the $\bm{x}_\text{prompt}$. This subsequently determines the score of targets (`positive' or `negative'). In detail:

\paragraph{Training strategy:} Let $\mathcal{M}: \mathcal{Y} \to \mathcal{V}$ be a mapping from target labels to individual words in a PLM's vocabulary. We refer to this mapping as \emph{verbalizers}. Then the input is converted to $\bm{x}_\text{prompt} = \mathcal{T}(\bm{x})$ by appending a \emph{pattern} and a \emph{mask token} to $\bm{x}$ so that it has the format of a masked language modeling input. Then, the classification task is converted to a MLM objective~\citep{tam2021improving, PET1}, and the PLM computes the probability of the label $y$ as: 

\begin{align}
p(y|\bm{x}) &= p(\text{[MASK]}= \mathcal{M}(y) |\bm{x}_\text{prompt}) \nonumber \\
&= \frac{\exp (\bm{W}_{\mathcal{M}(y)}^T \bm{h}_{\text{[MASK]}})}{\sum_{v'\in \mathcal{V}} \exp(\bm{W}^T_{v'}\bm{h}_\text{[MASK]})}, \label{eqn:pet_prob}
\end{align}
where $\bm{h}_\text{[MASK]}$ is the last hidden representation of the mask, and $\bm{W}_{v}$ shows the output embedding of the PLM for each verbalizer $v \in \mathcal{V}$. For many tasks, verbalizers have multiple tokens. \citet{PET2} extended \eqref{eqn:pet_prob} to multiple mask tokens by adding the maximum number of mask tokens $M$ needed to express the outputs (verbalizers) for a task. In that case, \citet{PET2} computes the probability of each class as the summation of the log probabilities of each token in the corresponding verbalizer, and then they add a hinge loss to ensure a margin between the correct verbalizer and the incorrect ones.

\paragraph{Inference strategy:} During inference, the model needs to select which verbalizer to use in the given context. \citet{PET2} predicts the verbalizer tokens in an autoregressive fashion. They first trim the number of mask tokens from $M$ to each candidate verbalizer's token length and compute the probability of each mask token. They then choose the predicted token with the highest probability and replace the corresponding mask token. Conditioning on this new token, the probabilities of the remaining mask positions are recomputed. They repeat this autoregressive decoding until they fill all mask positions. This inference strategy is very slow, as the number of forward passes increases with the number of classes and the number of verbalizer's tokens.

This formulation obtained impressive few-shot performance with PLMs. However, the success of this approach heavily relies on engineering handcrafted \emph{patterns} and \emph{verbalizers}. Coming up with suitable verbalizers and patterns can be difficult \citep{mishra2021crosstask, mishra2021reframing}. Additionally, the performance is sensitive to the wording of patterns \citep{zhao2021calibrate, perez2021true, PET1, jiang2020can} or to the chosen verbalizers \citep{webson2021prompt}.

\begin{figure}[!tp]
\centering 
\includegraphics[width=0.45\textwidth]{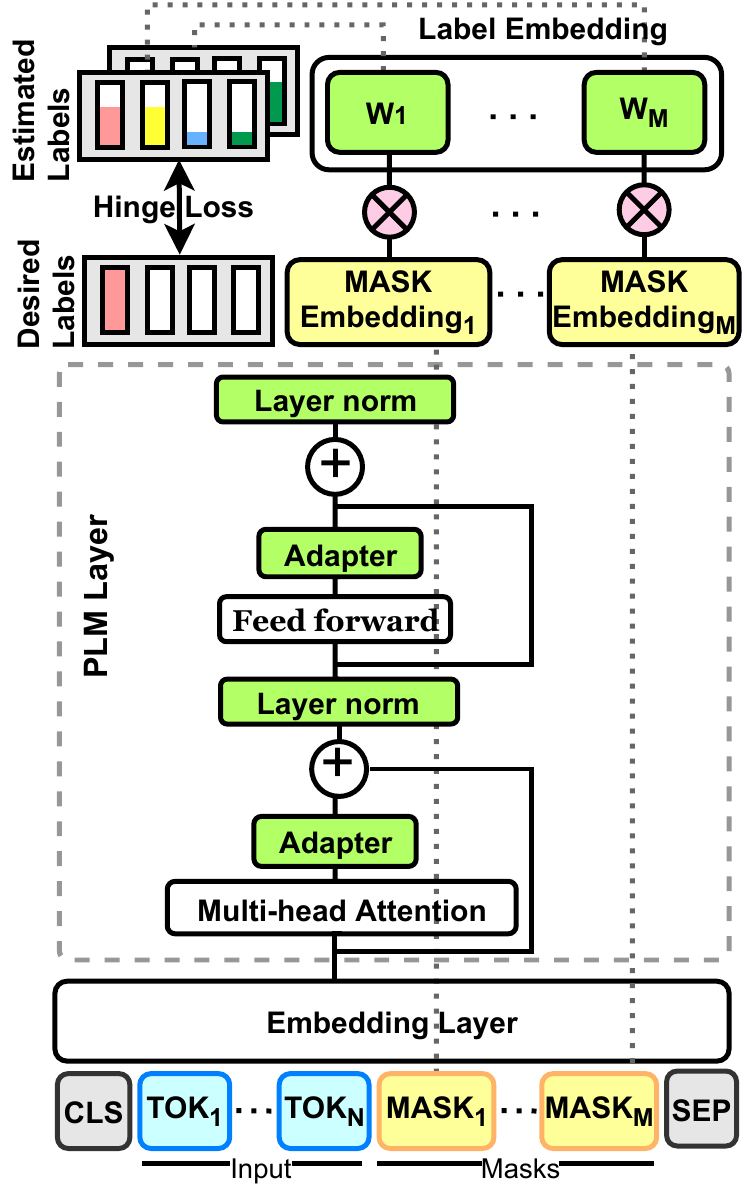}
\caption{We remove handcrafted patterns and verbalizers. We replace patterns using task-specific adapters and design label embeddings for the classes. We only train the green blocks (the label embeddings, adapters, and layer norms).}
\label{fig:perfect} 
\end{figure}

In addition, handcrafted verbalizers cause problems for efficient training:
a) they require updating the PLM embedding layer, causing large memory overhead; 
b) fine-tuning PLMs also requires a very small learning rate (usually $10^{-5}$), which slows down tuning the parameters of the verbalizers; 
c) modeling verbalizers as one of the tokens of the PLM vocabulary (perhaps unintentionally) impacts the input representation during tuning; 
d) verbalizers have variable token lengths, complicating the implementation in a vectorized format, thereby making it challenging to efficiently fine-tune PLMs.

\section{Method} 
We propose \perfect, a \emph{verbalizer and pattern free} few-shot learning method. 
 We design \perfect to be close to the pre-training phase, similar to the PET family of models~\citep{PET2, gao2020making}, while replacing handcrafted patterns and verbalizers with new components that are designed to describe the task and learn the labels. As shown in Figure \ref{fig:perfect}, we first convert each input $\bm{x}_\text{input}$ to its masked language modeling (MLM) input containing $M$ mask tokens \mask\footnote{We discuss the general case with inserting multiple masks; for some datasets this improves performance (\textsection\ref{sec:ablation}).}  with no added patterns, denoted as $\bm{x}_\text{masked} =\mathcal{T^{'}}(\bm{x}_\text{input})$.\footnote{We insert mask tokens after the input string in single-sentence benchmarks, and after the first sentence in the case of sentence-pair datasets and encode both sentences as a single input,  which we found to perform the best (Appendix \ref{app:mask_position}).}  \perfect then trains a classifier per-token and optimizes the average multi-class hinge loss over each mask position. 

 Three main components play a role in the success of \perfect: a) a pattern-free task description, where we use task-specific adapters to efficiently tell the model about the given task, replacing previously manually engineered patterns (\textsection\ref{sec:pattern_free}), b) multi-token label-embedding as an efficient mechanism to learn the label representations, removing manually designed verbalizers (\textsection \ref{sec:soft_verbalizers}). c) an efficient inference strategy building on top of the idea of prototypical networks \citep{snell2017prototypical} (\textsection\ref{sec:perfect_eval}), which replaces prior iterative autoregressive decoding methods~\citep{PET2}.  
 
 As shown in Figure~\ref{fig:perfect}, we fix the underlying PLM model and only optimize the new parameters that we add (green boxes). This includes the task-specific adapters to adapt the representations for a given task and the multi-token label representations. We detail each of these components below.

\subsection{Pattern-Free Task Description} \label{sec:pattern_free}
We use task-specific adapter layers to provide the model with learned, implicit task descriptions. Adapters additionally bring multiple other benefits: a) fine-tuning all weights of PLMs with millions or billions of parameters is sample-inefficient, and can be unstable in low-resource settings \citep{dodge2020fine}; adapters allow sample-efficient fine-tuning, by keeping the underlying PLM fixed, b) adapters reduce the storage and memory footprints (\textsection\ref{sec:efficiency}), c) they also increase stability and performance (\textsection\ref{sec:experiments}), making them an excellent choice for few-shot fine-tuning. To our knowledge, this is the first approach for using \emph{task-specific adapters} to effectively and efficiently remove patterns in few-shot learning. Experimental results in \textsection\ref{sec:experiments} show its effectiveness compared to handcrafted patterns and soft prompts~\citep{li2021prefix, lester2021power}.

\subsection{Multi-Token Label Embeddings}\label{sec:soft_verbalizers}
We freeze the weights of the PLM's embedding layer and introduce a separate label embedding $\bm{L}\in\mathbb{R}^{K\times M \times H}$, which is a multi-token label representation where $M$ is the number of tokens representing each label, $K$ indicates the number of classes, $H$ is the input hidden dimension. Using a fixed number of tokens $M$ for each label, versus variable-token length verbalizers used in prior work \citep{PET1, PET2} substantially simplifies the implementation and accelerates the training (\textsection\ref{sec:efficiency}).

\subsection{Training \perfect} 
As shown in Figure~\ref{fig:perfect}, we optimize label embeddings so that the PLM predicts the correct label, and optimize adapters to adapt the PLM for the given task. For label embeddings, \perfect trains a classifier per token and optimizes the average multi-class hinge loss over all mask positions. 
Given $\bm{x}_\text{masked}$, let $\bm{h}_{\mask_i}$ be the embedding of its $i$-th mask token from the last layer of the PLM encoder. Additionally, let $f(.): \mathbb{R}^H \to \mathbb{R}^{K}$ be a per-token classifier that computes the predictions by multiplying the mask token embedding with its corresponding label embedding. Formally defined as:
\begin{align}
\bm{t_i} = f(\bm{h}_{\mask_i}) = \bm{L_i}^T \bm{h}_{\mask_i},
\label{eqn:label-embeddings} \nonumber 
\end{align}
where $\bm{L_i}\in\mathbb{R}^{K\times H}$ shows the label embedding for the $i$-th mask position. Then, for each mask position, we optimize a multi-class hinge loss between their scores $\bm{t_i}$ and labels. Formally defined as: 
\begin{align}
\mathcal{L}(\bm{x}, y, i) = \frac{\sum_{k=1, k\neq y}^K \max(0, m-\bm{t_{iy}}+\bm{t_{ik}}) }{K}, \nonumber 
\end{align}
where $\bm{t_{ik}}$ shows the $k$-th element of $\bm{t_i}$, representing the score corresponding to class $k$, and $m$ is the margin, which we fix to the default value of $m=1$. Then, the final loss is computed by averaging the loss over all mask tokens and training samples:
\begin{align}
\mathcal{L} = \frac{1}{M|\mathcal{D}|}\sum_{(\bm{x},y)\in\mathcal{D}}\sum_{i=1}^M\mathcal{L}(\bm{x}, y, i) 
\\[-5ex]\nonumber
\label{eqn:total_loss}
\end{align}

\subsection{Inference with \perfect}\label{sec:perfect_eval} 
During evaluation, instead of relying on the prior iterative autoregressive decoding schemes~\citep{PET2}, we classify a query point by finding the nearest class prototype to the mask token embeddings: 
\begin{align}
y = \argmax_{y \in \mathcal{Y}}\hspace{0.1em} \max_{i \in \{1, \dots, M\}} \left( \hspace{0.1em}{\exp^{-d(\bm{h_{i}^q}, \bm{c_{iy}})}} \right), 
\\[-5ex]\nonumber
\end{align}
where $d$ is squared euclidean distance,\footnote{We also tried with cosine similarity but found a slight improvement with squared Euclidean distance \citep{snell2017prototypical}.} $\bm{h_{i}^q}$ indicates the embedding of the $i$-th mask position for the query sample $q$, and  $\bm{c_{iy}} \in \mathbb{R}^D$ is the prototype representation of the $i$-th mask token with class label $y$, i.e., the mean embedding of $i$-th mask position in all training samples with label $y$:
\begin{align}
\bm{c_{iy}} = \frac{1}{|\mathcal{D}_y|} \sum_{b \hspace{0.01em} \in \mathcal{D}_y} \bm{h_i^b},
\label{eqn:centroids} 
\end{align} 
where $\bm{h_i^b}$ shows the embedding of $i$-th mask position for training sample $b$, and $\mathcal{D}_y$ is the training instances with class $y$. This strategy closely follows prototypical networks \citep{snell2017prototypical}, but applied across multiple tokens. We choose this form of inference because prototypical networks are known to be sample efficient and robust \citep{snell2017prototypical}, and because it substantially speeds up evaluation compared to prior methods (\textsection\ref{sec:efficiency}).

 \begin{table*}[tp]
\centering 
\vspace{-1em} 
\begin{adjustbox}{max width=1.0\linewidth}
\begin{tabular}{l@{\hskip 0.02in}|@{\hskip 0.02in}l@{\hskip 0.1in}l@{\hskip 0.1in}l@{\hskip 0.1in}l@{\hskip 0.1in}l@{\hskip 0.1in}l@{\hskip 0.02in}|@{\hskip 0.02in}l}
\toprule 
\textbf{Method} & \textbf{SST-2} & \textbf{CR} & \textbf{MR} &    \textbf{SST-5} &   \textbf{Subj} &   \textbf{TREC} &     \textbf{Avg} \\
\midrule 
\rowcolor{gray!20}\multicolumn{8}{c}{\it \textbf{Single-Sentence Benchmarks}}\\
\midrule 
\finetune & 81.4\smaller{/70.0/4.0}&80.1\smaller{/72.9/4.1}&77.7\smaller{/66.8/4.6}&39.2\smaller{/34.3/2.5}&\textbf{90.2}\smaller{/84.1/\textbf{1.8}}&87.6\smaller{/75.8/3.7}&76.0\smaller{/67.3/3.4}\\ 
\pet-Average & 89.7\smaller{/81.0/2.4}&88.4\smaller{/68.8/3.0}&85.9\smaller{/79.0/2.1}&45.9\smaller{/40.3/\textbf{2.4}}&88.1\smaller{/79.6/2.4}&85.0\smaller{/70.6/4.5}&80.5\smaller{/69.9/2.8}\\ 
\pet-Best & 89.1\smaller{/81.0/2.6}&88.8\smaller{/85.8/1.9}&\textbf{86.4\smaller{/82.0/1.6}}&\textbf{46.0}\smaller{/41.2/2.4}&88.7\smaller{/\textbf{84.6/1.8}}&85.8\smaller{/70.6/4.4}&80.8\smaller{/74.2/2.4}\\ 
\citet{logan2021cutting} & 89.8\smaller{/84.1/1.7}&89.9\smaller{/87.2/1.1}&84.9\smaller{/76.2/3.2}&45.7\textbf{\smaller{/41.6/2.3}}&81.8\small{/73.5/4.0}&84.7\smaller{/81.8/1.6}&79.5\smaller{/74.1/2.3} \\ 
\midrule 
\perfect-rand & 90.7\textbf{\smaller{/88.2/1.2}}&90.0\smaller{/85.5/1.4}&86.3\smaller{/81.4/\textbf{1.6}}&42.7\smaller{/35.1/2.9}&89.1\smaller{/82.8/2.1}&\textbf{90.6}\smaller{/81.6/3.2}&\textbf{81.6}\smaller{\textbf{/75.8/2.1}}\\
\midrule 
\rowcolor{gray!5}\multicolumn{8}{c}{\it \textbf{Ablation}}\\
\midrule
\perfect-init & \textbf{90.9}\smaller{/87.6/1.5}&89.7\smaller{/87.4/1.2}&85.4\smaller{/75.8/3.3}&42.8\smaller{/35.9/3.5}&87.6\smaller{/81.6/2.8}&90.4\smaller{/\textbf{86.6/1.8}}&81.1\smaller{/\textbf{75.8}/2.4}\\ 
prompt+mte & 70.6\smaller{/56.0/8.3}&71.0\smaller{/55.8/8.2}&66.6\smaller{/49.6/7.3}&32.2\smaller{/26.5/3.2}&82.7\smaller{/69.6/3.9}&79.6\smaller{/66.8/6.5}&67.1\smaller{/54.0/6.2}\\ 
bitfit+mte & 89.5\smaller{/81.7/3.0}&\textbf{90.1\smaller{/87.8/1.0}}&85.6\smaller{/80.5/1.9}&42.3\smaller{/36.8/3.3}&89.1\smaller{/82.4/2.4}&90.4\smaller{/85.0/1.4} & 81.2/\smaller{75.7/2.2}\\ 
\toprule
\textbf{Method} & \textbf{CB} & \textbf{RTE} & \textbf{QNLI} &    \textbf{MRPC} &   \textbf{QQP} &   \textbf{WiC} &  \textbf{Avg} \\
\midrule  
\rowcolor{gray!20}\multicolumn{8}{c}{\it \textbf{Sentence-Pair Benchmarks}}\\
\midrule 
\finetune & 72.9\smaller{/67.9/\textbf{2.5}}&56.8\smaller{/50.2/\textbf{3.5}}&62.7\smaller{/51.4/7.0}&\textbf{70.1\smaller{/62.7/4.7}}&65.0\smaller{/59.8/3.6}&52.4\smaller{/46.1/3.7}&63.3\smaller{/56.4/4.2}\\
\pet-Average &86.9\smaller{/73.2/5.1}&60.1\smaller{/49.5/4.7}&66.5\smaller{/55.7/6.2}&62.1\smaller{/38.2/6.8}&63.4\smaller{/44.7/7.9}&51.0\smaller{/46.1/2.6} & 65.0/\smaller{51.2/5.6}\\ 
\pet-Best    &90.0\smaller{/78.6/3.9}&62.3\smaller{/51.3/4.5}&70.5\smaller{/57.9/6.4}&63.4\smaller{/49.3/6.5}&70.7\smaller{/55.2/5.8}& 51.6\smaller{/47.2/2.3} & 68.1/\smaller{56.6/4.9}\\ 
\citet{logan2021cutting} & 91.0\smaller{/87.5/2.7}&\textbf{64.4\smaller{/58.5}}\smaller{/3.9}&71.2\smaller{/\textbf{66.5/2.6}}&63.9\smaller{/53.7/5.3}&70.4\smaller{/62.7/\textbf{3.4}}&52.4\smaller{\textbf{/48.4/1.8}}&68.9\smaller{/\textbf{62.9/3.3}}\\
\midrule 
\perfect-rand & \textbf{90.3}\smaller{/\textbf{83.9}/3.5}&60.4\smaller{/53.1/4.7}&\textbf{74.1}\smaller{/60.3/4.6}&67.8\smaller{/54.7/5.7}&\textbf{71.2}\smaller{/64.2/3.5}&\textbf{53.8}\smaller{/47.0/3.0}&\textbf{69.6}\smaller{/60.5/4.2}\\
\midrule 
\rowcolor{gray!5}\multicolumn{8}{c}{\it \textbf{Ablation}}\\
\midrule
\perfect-init & 87.9\smaller{/75.0/4.9}&60.7\smaller{/52.7/4.5}&72.8\smaller{/56.7/6.8}&65.9\smaller{/56.6/6.0}&71.1\smaller{/\textbf{65.6}/3.5}&51.7\smaller{/46.6/2.8}&68.4\smaller{/58.9/4.8}\\
prompt+mte & 73.0\smaller{/62.5/6.1}&56.9\smaller{/50.7/4.1}&55.4\smaller{/50.2/4.6}&60.0\smaller{/51.5/5.8}&54.3\smaller{/46.2/5.6}&51.3\smaller{/46.7/2.8}&58.5\smaller{/51.3/4.8}\\ 
bitfit+mte & 89.6\smaller{/82.1/4.3}&61.3\smaller{/53.8/5.2}&70.6\smaller{/51.9/5.9}&68.5\smaller{/57.4/5.1}&69.4\smaller{/63.0/3.9}&52.9\smaller{/47.8/2.7}&68.7/\smaller{59.3/4.5} \\
\bottomrule
\end{tabular}
\end{adjustbox}
\caption{Performance of all methods on single-sentence and sentence-pair benchmarks. We report average/worst-case accuracy/standard deviation. \perfect obtains the state-of-the-art results. Bold fonts indicate the best results.}
\label{tab:results} 
\end{table*}

\section{Experiments} \label{sec:experiments}
We conduct extensive experiments on a variety of NLP datasets to evaluate the performance of \perfect and compare it with state-of-the-art few-shot learning.

\paragraph{Datasets:}We consider 7 tasks and 12 datasets:~1) the sentiment analysis datasets SST-2~\citep{socher2013recursive}, SST-5~\citep{socher2013recursive}, MR~\citep{pang2005seeing}, and CR~\citep{hu2004mining},~2) the subjectivity classification dataset SUBJ~\citep{pang2004sentimental},~3) the question classification dataset TREC~\citep{voorhees2000building},~4) the natural language inference datasets CB~\citep{de2019commitmentbank} and RTE~\citep{wang2019superglue}, 5) the question answering dataset QNLI~\citep{rajpurkar2016squad},~6) the word sense disambiguation dataset WiC~\citep{pilehvar2019wic},~7) the paraphrase detection datasets MRPC~\citep{dolan2005automatically} and QQP.\footnote{\url{https://quoradata.quora.com/}} See datasets statistics in Appendix~\ref{app:details}.

For MR, CR, SST-5, SUBJ, and TREC, we test on the original test sets, while for other datasets, since test sets are not publicly available, we test on the original validation set. We sample 16 instances per label from the training set to form training and validation sets. 

\paragraph{Baselines} We compare with the state-of-the-art few-shot learning of PET and fine-tuning:

\textbf{\pet}~\citep{PET1, PET2}$\:$ is the state-of-the-art few-shot learning method that employs carefully crafted verbalizers and patterns. We report the best (PET-best) and average (PET-average) results among all patterns and verbalizers.\footnote{For a controlled study, we use the MLM variant shown in \eqref{eqn:pet_prob}, which has been shown to perform the best \citep{tam2021improving}.} 

\textbf{\finetune}$\:$ The standard fine-tuning \citep{devlin-etal-2019-bert}, with adding a classifier on top of the [CLS] token and fine-tuning all parameters. 

\paragraph{Our method} We study the performance of \perfect and perform an extensive ablation study to show the effectiveness of our design choices:

\textbf{\perfect-rand}$\:$ We randomly initialize the label embedding $\bm{L}$ from a normal distribution $\mathcal{N} (0, \sigma)$ with $\sigma=10^{-4}$ (chosen based on validation performance, see Appendix \ref{app:sigma}) \emph{without relying on any handcrafted patterns and verbalizers}. As an ablation, we study the following two variants:

\textbf{\perfect-init}$\:$ We initialize the label embedding with the token embeddings of manually designed verbalizers in the PLM's vocabulary to study the impact of engineered verbalizers.

\textbf{prompt+mte}$\:$ To compare the impact of adapters versus soft prompt-tuning for few-shot learning, we append trainable continuous prompt embeddings to the input \citep{lester2021power}. Then we only tune the soft prompt and multi-token label embeddings (mte).

\textbf{bitfit+mte}$\:$ Following~\citet{cai2020tinytl} and~\citet{ravfogel2021bitfit}, we tune biases as an alternative to adapters. We additionally tune multi-token label embeddings.

\textbf{\citet{logan2021cutting}} Following ~\citet{logan2021cutting}, we remove patterns and tune the biases in the PET.

\paragraph{Experimental details:} We use the RoBERTa large model \citep{liu2019roberta} (355M parameters) as the underlying PLM for all methods. We use the HuggingFace PyTorch implementation \citep{wolf-etal2020transformers}. For the baselines, we used the carefully manually designed patterns and verbalizers in \citet{gao2020making}, \citet{min2021noisy}, and \citet{PET2} (usually 5 different options per datasets; see Appendix \ref{app:prompts}). 

We evaluate all methods using 5 different random samples to create the training/validation sets and 4 different random seeds for training. Therefore, for PET-average, we report the results on 20 x 5 (number of patterns and verbalizers) = 100  runs, while for PET-best and our method, we report the results over 20 runs. The variance in few-shot learning methods is usually high~\citep{perez2021true, zhao2021calibrate, lu2021fantastically}. Therefore, we report average, worst-case performance, and standard deviation across all runs, where the last two values can be important for risk-sensitive applications \citep{asri2016using}. 

\subsection{Experimental Results}
Table \ref{tab:results} shows the performance of all methods. \perfect obtains state-of-the-art results, improving the performance compared to \pet-average by +1.1 and +4.6 points for single-sentence and sentence-pair datasets respectively. It even outperforms \pet-best, where we report the best performance of \pet across multiple manually engineered patterns and verbalizers. Moreover, \perfect generally improves the minimum performance and reduces standard deviation substantially.  Finally, \perfect is also significantly more efficient: reducing the training and inference time, memory usage, and storage costs (see \textsection \ref{sec:efficiency}). 

\pet-best improves the results over \pet-average showing that \pet is unstable to the choice of patterns and verbalizers; this difference is more severe for sentence-pair benchmarks. This might be because the position of the mask highly impacts the results, and the patterns used for sentence-pair datasets in \citet{PET2} exploits this variation by putting the mask in multiple locations (see Appendix~\ref{app:prompts}). 

Removing patterns and tuning biases in \citet{logan2021cutting} is not expressive enough and performs substantially worse than \perfect on average.

As an ablation, even if we initialize the label embedding with handcrafted verbalizers in \perfect-init, it consistently obtains lower performance, demonstrating that \perfect is able to obtain state-of-the-art performance with learning from \emph{pure random initialization}. We argue that initializing randomly close to zero (with low variance $\sigma=10^{-4}$), as done in our case, slightly improves performance, which perhaps is not satisfied when initializing from the manually engineered verbalizers (see Appendix \ref{app:sigma}).

As a second ablation, when learning patterns with optimizing soft prompts in prompt+mte, we observe high sensitivity to learning rate, as also confirmed in~\citet{li2021prefix} and \citet{mahabadi2021compacter}. We experimented with multiple learning rates but performance consistently lags behind \perfect-rand. This can be explained by the low flexibility of such methods as all the information regarding specifying patterns needs to be contained in the prefixes. As a result, the method only allows limited interaction with the rest of the model parameters, and obtaining good performance requires very large models~\citep{lester2021power}. In addition, increasing the sequence length leads to memory overhead \citep{mahabadi2021compacter}, and the number of prompt tokens is capped by the number of tokens that can fit in the maximum input length, which can be a limitation for tasks requiring large contexts.

As a third ablation, tuning biases with optimizing soft prompts in bitfit+mte obtains lower performance compared to \perfect, showing that adapters are a better alternative compared to tuning biases to learn task descriptions for few-shot learning. 

We include more ablation results on design choices of \perfect in Appendix~\ref{app:ablation}.

\begin{table}[tp] 
\centering 
\begin{adjustbox}{max width=1.0\linewidth} 
\begin{tabular}{l@{\hskip 0.05in}l@{\hskip 0.05in}l@{\hskip 0.05in}l}
\toprule 
\textbf{Metric} & \textbf{\pet} & \textbf{\perfect}  & $\bm{\Delta \%}$\\
\toprule 
Trained params (M) & 355.41 & \textbf{3.28} & \textbf{-99.08\%} \\ 
Peak memory (GB) & 20.93 & \textbf{16.34} &\textbf{-21.93\%}\\ 
Training time (min) & 23.42 & \textbf{0.65} & \textbf{-97.22\%}\\
\hspace{1.5em} + PET in batch & 0.94 & \textbf{0.65} & \textbf{-30.85\%}\\
Inference time (min) & 9.57 &  \textbf{0.31} & \textbf{-96.76\%}\\ 
\bottomrule
\end{tabular}
\end{adjustbox}
\caption{Percentage of trained parameters, average peak memory, training, and inference time. $\bm{\Delta \%}$ is the relative difference with respect to \pet. Lower is better.}
\label{tab:performance}
\end{table}

\subsection{Efficiency Evaluation} \label{sec:efficiency}
In this section, we compare the efficiency of \perfect with the state-of-the-art few-shot learning method, PET. To this end, we train all methods for ten epochs on the 500-sampled QNLI dataset. We select the largest batch size for each method that fits a fixed budget of the GPU memory (40 GB).  

Due to the auto-regressive inference strategy of PET~\citep{PET2}, all prior work implemented it with a batch size of 1 \citep{perez2021true, PET2, tam2021improving}. Additionally, since PET deals with verbalizers of variable lengths, it is hard to implement their training phase in batch mode. We specifically choose QNLI to have verbalizers of the same length and enable batching for comparison purposes (referred to as \emph{PET in batch}). However, verbalizers are still not of fixed-length for most other tasks, and this speed-up does not apply generally to PET.

In Table~\ref{tab:performance}, for each method we report the percentage of trained parameters, memory usage, training time, and inference time. 
\jh{redundant: In addition to} \perfect reduces the number of trained parameters, and therefore the storage requirement, by 99.08\%. It additionally reduces the memory requirement by 21.93\% compared to \pet. \perfect speeds up training substantially, by 97.22\% relative to the original \pet's implementation, and 30.85\% to our implementation of PET. This is because adapter-based tuning saves on memory and allows training with larger batch sizes.  In addition, \perfect is significantly faster during inference time (96.76\% less inference time relative to \pet). 

Note that although prompt+mte and bitfit+mte can also reduce the storage costs, by having 0.02M and 0.32 M trainable parameters respectively, they are not expressive enough to learn task descriptions, and their performance substantially lags behind \perfect (see Table~\ref{tab:results}).

Overall, given the size of PLMs with millions and billions of parameters~\citep{liu2019roberta, raffel2020exploring}, efficient few-shot learning methods are of paramount importance for practical applications. \perfect not only outperforms the state-of-the-art in terms of accuracy and generally improves the stability (Table~\ref{tab:results}), but also is significantly more efficient in runtime, storage, and memory.

\begin{table}[tp]
\centering 
\begin{tabular}{l@{\hskip 0.05in}|@{\hskip 0.05in}l@{\hskip 0.15in}l}
\toprule 
\textbf{Dataset} & \textbf{\pet-Average} & \textbf{Pattern-Free} \\
\midrule 
SST-2 & 89.7/81.0/2.4 & \textbf{90.5/87.8/1.2} \\
CR    &88.4/68.8/3.0  & \textbf{89.8/87.0/1.4}\\
MR    &85.9/79.0/2.1  & \textbf{86.4/83.0/1.8} \\
SST-5 &\textbf{45.9/40.3/2.4} & 44.8/40.0/2.4\\
SUBJ & \textbf{88.1/79.6/2.4} & 85.3/74.7/3.8 \\
TREC &85.0/70.6/4.5   &\textbf{87.9/84.6/1.8}\\
CB &86.9/73.2/5.1& \textbf{93.0/89.3/1.9}\\
RTE & 60.1/49.5/4.7 & \textbf{63.7/56.3/4.1} \\ 
QNLI &66.5/55.7/6.2 & \textbf{71.3/65.8/2.5} \\ 
MRPC &62.1/38.2/6.8& \textbf{66.0/54.4/5.6} \\ 
QQP &63.4/44.7/7.9&\textbf{71.8/64.3/3.7}\\ 
WiC &51.0/46.1/2.6 & \textbf{53.7/50.3/2.0} \\
\midrule 
Avg & 72.8/60.6/4.2 &  \textbf{75.4/69.8/2.7}\\ 
\bottomrule
\end{tabular}
\caption{Average performance of \emph{PET} with five different patterns vs. \emph{Pattern-Free} that replaces handcrafted patterns with task-specific adapters. We report the average/worst-case performance/and the standard deviation.}
\label{tab:remove_patterns} 
\end{table} 
\subsection{Analysis}\label{sec:analysis} 
\paragraph{Can task-specific adapters replace manually engineered patterns?} \perfect is a pattern-free approach and employs adapters to provide the PLMs with task descriptions implicitly. In this section, we study the contribution of replacing manual patterns with adapters in isolation without considering our other contributions in representing labels, training, and inference. 
In PET~\citep{PET1, PET2}, we replace the handcrafted patterns with task-specific adapters (\emph{Pattern-Free}) while keeping the verbalizers and the training and inference intact\footnote{Since we don't have patterns, in the case of multiple sets of verbalizers, we use the first set of verbalizers as a random choice.} and train it with a similar setup as in \textsection\ref{sec:experiments}. Table \ref{tab:remove_patterns} shows the results. While PET is very sensitive to the choice of prompts, adapters provide an efficient alternative to learn patterns robustly by improving the performance (average and worst-case) and reducing the standard deviation. This finding demonstrates that task-specific adapters can effectively replace manually engineered prompts. Additionally, they also save on the training budget by at least $1/\text{number of patterns}$ (normally 1/5) by not requiring running the method for different choices of patterns, and by freezing most parameters, this saves on memory and offers additional speed-up. 

\subsubsection{Ablation Study} \label{sec:ablation}
\paragraph{Impact of Removing Adapters} To study the impact of adapters in learning patterns, we remove adapters, while keeping the label embedding. Handcrafted patterns are not included and we tune all parameters of the model. Table \ref{tab:remove_adapters} shows the results. Adding adapters for learning patterns contributes to the performance by improving the average performance, and making the model robust by improving the minimum performance and reducing the standard deviation. This is because training PLMs with millions of parameters is sample-inefficient and unstable on resource-limited datasets~\citep{dodge2020fine, zhang2020revisiting,mosbach2021on}. However, by using adapters, we substantially reduce the number of trainable parameters, allowing the model to be better tuned in a few-shot setting.

\begin{table}[tp]
\centering 
\begin{tabular}{l|@{\hskip 0.05in}l@{\hskip 0.15in}l}
\toprule 
\textbf{Dataset} & \textbf{\perfect} & \textbf{-Adapters} \\
\midrule 
SST-2 & \textbf{90.7/88.2/1.2} &88.2/81.9/2.3\\
CR    & \textbf{90.0/85.5/1.4} &89.2/83.1/1.7\\
MR    & \textbf{86.3/81.4/1.6} &82.5/78.2/2.5 \\
SST-5 & \textbf{42.7/35.1/2.9} &40.6/33.6/3.3\\
SUBJ  & 89.1/82.8/2.1 & \textbf{89.7/85.0/1.9}\\
TREC  & \textbf{90.6/81.6/3.2} &89.8/74.2/4.3\\
CB    &\textbf{90.3/83.9}/3.5 &89.6\textbf{/83.9/2.8}\\
RTE   & 60.4/53.1/\textbf{4.7}&\textbf{61.7}\textbf{/53.8}/5.1\\ 
QNLI  & \textbf{74.1/60.3/4.6}&73.2/56.3/5.8\\ 
MRPC  & 67.8\textbf{/54.7/5.7}&\textbf{68.0}/54.2/6.1\\ 
QQP   &\textbf{71.2/64.2/3.5} &71.0/62.0/3.7\\ 
WiC   &\textbf{53.8/47.0/3.0} &52.5/46.9/3.0\\
\midrule 
Avg & \textbf{75.6/68.1/3.1} & 74.7/66.1/3.5 \\ 
\bottomrule
\end{tabular}
\caption{Performance of \emph{\perfect} w/o adapters, \emph{-Adapters}. We report the average performance/worst-case performance/and the standard deviation.}
\label{tab:remove_adapters} 
\end{table}

\paragraph{Impact of the number of masks} In Table~\ref{tab:results}, to compare our design with PET in isolation, we fixed the number of mask tokens as the maximum number inserted by PET. In table \ref{tab:num_masks}, we study the impact of varying the number of inserted mask tokens for a random selection of six tasks. For most tasks, having two mask tokens performs the best, while for MR and RTE, having one, and for MRPC, inserting ten masks improves the results substantially. The number of required masks might be correlated with the difficulty of the task. \perfect is designed to be general, enabling having multiple mask tokens.

\section{Related Work} \label{sec:related_work}
\paragraph{Adapter Layers:} \citet{karimi2021parameter-efficient} and \citet{ustun2020udapter} proposed to generate adapters' weights using hypernetworks \citep{Ha2017hypernetworks}, where \citet{karimi2021parameter-efficient} proposed to share a small hypernetwork to generate conditional adapter weights efficiently for each transformer layer and task. \citet{mahabadi2021compacter} proposed compacter layers by building on top of ideas of parameterized hyper-complex layers \citep{zhang2021beyond} and low-rank methods \citep{li2018measuring, aghajanyan2020intrinsic}, as an efficient fine-tuning method for PLMs. We are the first to employ adapters to replace handcrafted patterns for few-shot learning.

\begin{table}[tp]
\centering
\begin{adjustbox}{max width=1.0\linewidth}
\begin{tabular}{lllll}
\toprule 
\textbf{Datasets} & \textbf{1} & \textbf{2} & \textbf{5} & \textbf{10}\\
\toprule 
CR & 90.1 &  \textbf{90.2} & 89.0 & 87.8 \\
MR & \textbf{86.9} &  86.1 & 85.4&85.6 \\ 
MRPC & 67.4 &68.2 & 70.1 & \textbf{72.3}\\
QNLI &73.7  & \textbf{73.9} &73.0 &65.1 \\
RTE & \textbf{60.0} &  57.3 &56.2&56.0 \\ 
TREC & 90.0 & \textbf{90.9}  &88.9&   88.8\\
\midrule 
Avg &\textbf{78.0}& 77.8& 77.1& 75.9\\
\bottomrule
\end{tabular}
\end{adjustbox}
\caption{Test performance for the varying number of mask tokens. Bold fonts indicate the best results in each row.}
\label{tab:num_masks}
\end{table}
\paragraph{Few-shot Learning with PLMs:} \citet{le2021many} showed that prompting provides substantial improvements compared to fine-tuning, especially in low-resource settings. Subsequently, researchers continuously tried to address the challenges of manually engineered patterns and verbalizers: a) Learning the patterns in a continuous space~\citep{li2021prefix, qin2021learning, lester2021power}, while freezing PLM for efficiency, has the problem that, in most cases, such an approach only works with very large scale PLMs \citep{lester2021power}, and lags behind full fine-tuning in a general setting, while being inefficient and not as effective compared to adapters~\citep{mahabadi2021compacter}. b) Optimizing patterns in a discrete space \citep{shin2020eliciting, jiang2020can, gao2020making} has the problem that such methods are computationally costly.  c) Automatically finding verbalizers in a discrete way \cite{schick2020automatically, PET1} is computationally expensive and does not perform as well as manually designed ones. d) Removing manually designed patterns~\citep{logan2021cutting} substantially lags behind the expert-designed ones. Our proposed method, \perfect, does not rely on any handcrafted patterns and verbalizers.

\section{Conclusion}
We proposed \perfect, a simple and efficient method for few-shot learning with pre-trained language models without relying on handcrafted patterns and verbalizers. \perfect employs task-specific adapters to learn task descriptions implicitly, replacing previous handcrafted patterns, and a continuous multi-token label embedding to represent the output classes. Through extensive experiments over 12 NLP benchmarks, we demonstrate that \perfect, despite being far simpler and more efficient than recent few-shot learning methods, produces state-of-the-art results. Overall, the simplicity and effectiveness of \perfect make it a promising approach for few-shot learning with PLMs.

\section*{Acknowledgements}
The authors would like to thank Sebastian Ruder and Marius Mosbach for their comments  on drafts of this paper. This research was partly supported by the Swiss National Science Foundation under grant number 200021\_178862.

\bibliography{anthology}
\bibliographystyle{acl_natbib}

\appendix
\clearpage

\section{Experimental Details} \label{app:details} 

\paragraph{Datasets}Table \ref{tab:datasets} shows the stastistics of the datasets used. We download SST-2, MR, CR, SST-5, and SUBJ from \citet{gao2020making}, while the rest of the datasets are downloaded from the HuggingFace Datasets library \citep{lhoest-etal-2021-datasets,quentin_lhoest_2021_5639822}. RTE, CB, WiC datasets are from SuperGLUE benchmark \citep{wang2019superglue}, while QQP, MRPC and QNLI are from GLUE benchmark \citep{wang2018glue} with Creative Commons license (CC BY 4.0). RTE~\citep{wang2019superglue} is a combination of data from RTE1~\citep{dagan2005pascal}, RTE2~\citep{rte2}, RTE3~\citep{giampiccolo-etal-2007-third}, and RTE5~\citep{Bentivogli09thefifth}. For WiC \citep{pilehvar2019wic} sentences are selected from VerbNet~\citep{schuler2005verbnet}, WordNet~\citep{miller1995wordnet}, and Wiktionary. 

\begin{table}[tp]
\centering 
\begin{adjustbox}{max width=1\linewidth}
\begin{tabular}{l@{\hskip 0.05in}l@{\hskip 0.05in}l@{\hskip 0.05in}l@{\hskip 0.05in}l}
\toprule 
\textbf{Dataset} & \textbf{Task} & \textbf{\#Train} & \textbf{\#Test} & \textbf{K} \\
\midrule 
\rowcolor{gray!20}\multicolumn{5}{c}{\it \textbf{Single-Sentence Benchmarks}}\\
\midrule 
MR    &Sentiment analysis &8662 &2000 & 2\\ 
CR    &Sentiment analysis &1774 & 2000& 2 \\ 
SST-2 &Sentiment analysis & 6920& 872& 2 \\
SST-5 &Sentiment analysis &8544 & 2210  &5 \\
SUBJ  &Subjectivity classification &8000 &2000 &2 \\
TREC  &Question classification  &5452 & 500 &6 \\
\midrule 
\rowcolor{gray!20}\multicolumn{5}{c}{\it \textbf{Sentence-Pair Benchmarks}}\\
\midrule 
CB & Natural language inference &250&56& 3\\
RTE & Natural language inference &2490&277& 2 \\
WiC & Word sense disambiguation &5428&638& 2\\ 
MRPC &Paraphrase detection& 3668 & 408 & 2\\
QNLI &Question answering&  104743 & 5463 & 2\\
QQP &Paraphrase detection&363846 &40430 & 2\\
\bottomrule
\end{tabular}
\end{adjustbox}
\caption{Statistics of datasets used in this work. We sample $N \times |\mathcal{Y}|$ instances (with multiple seeds) from the original training set to form the few-shot training and validation sets. The test column shows the size of the test set.}
\label{tab:datasets} 
\end{table}

\paragraph{Computing infrastructure}We run all the experiments on one \textsc{NVIDIA A100} with 40G of memory.

\paragraph{Training hyper-parameters} We set the maximum sequence length based on the recommended values in the HuggingFace repository \citep{wolf-etal2020transformers} and prior work \citep{min2021noisy, PET2}, i.e., we set it to 256 for SUBJ, CR, CB, RTE, and WiC, and 128 for other datasets. For all methods, we use a batch size of 32. For \finetune and \pet, we use the default learning rate of $10^{-5}$, while for our method, as required by adapter-based methods \citep{mahabadi2021compacter}, we set the learning rate to a higher value of $10^{-4}$.\footnote{We have also tried to tune the baselines with the learning rate of $10^{-4}$ but it performed worst.} Through all experiments, we fix the adapter bottleneck size to 64. Following \citet{Pfeiffer2021adapterfusion}, we experimented with keeping one of the adapters in each layer for better training efficiency and found keeping the adapter after the feed-forward module in each layer to perform the best. For tuning label embedding, we use the learning rate of $\{10^{-1}, 10^{-2}, 10^{-3}, 10^{-4}, 10^{-5}\}$ and choose
the one obtaining the highest validation performance. For \perfect-prompt, we tune the continuous prompt for learning rate of $\{10^{-1}, 10^{-2}, 10^{-3}\}$.\footnote{We also tried tuning prompts with learning rates of $\{10^{-4}, 10^{-5}\}$ but it performed worst, as also observed in prior work \citep{mahabadi2021compacter, min2021noisy}.}Following \citet{lester2021power}, for \perfect-prompt, we set the number of prompt tokens to 20, and initialize them with a random subset of the top 5000 token's embedding of the PLM. We train all methods for 6000 steps. Based on our results, this is sufficient to allow the models to converge. We save a checkpoint every 100 steps for all methods and report the results for the hyper-parameters performing the best on the
validation set for each task.

\section{Choice of Patterns and Verbalizers} \label{app:prompts}
For SST-2, MR, CR, SST-5, and TREC, we used 4 different patterns and verbalizers from \citet{gao2020making}. For CB, WiC, RTE datasets, we used the designed patterns and verbalizers in \citet{PET2}. For QQP, MRPC, and QNLI, we wrote the patterns and verbalizers inspired by the ones in \citet{PET2}. The used patterns and verbalizers are as follows:

\begin{itemize}
    \item For sentiment analysis tasks (\textbf{MR}, \textbf{CR}, \textbf{SST-2}, \textbf{SST-5}), given a sentence $s$:
    \begin{tcolorbox}[width=0.9\columnwidth, colback=white!95!black]\small $s$ A <MASK> one.  \end{tcolorbox} \begin{tcolorbox}[width=0.9\columnwidth, colback=white!95!black]\small $s$ It was <MASK>. \end{tcolorbox} \begin{tcolorbox}[width=0.9\columnwidth, colback=white!95!black]\small $s$ All in all <MASK>.\end{tcolorbox} \begin{tcolorbox}[width=0.9\columnwidth, colback=white!95!black]\small $s$ A <MASK> piece. \end{tcolorbox}
    with "great" as a verbalizer for positive, "terrible" for negative. In case of SST-5 with five labels, we expand it to "great", "good", "okay", "bad", and "terrible".

    \item For \textbf{SUBJ}, given a sentence $s$:
    \begin{tcolorbox}[width=0.9\columnwidth, colback=white!95!black]\small $s$ This is <MASK>.  \end{tcolorbox} \begin{tcolorbox}[width=0.9\columnwidth, colback=white!95!black]\small $s$ It's all <MASK>. \end{tcolorbox} \begin{tcolorbox}[width=0.9\columnwidth, colback=white!95!black]\small $s$ It's <MASK>.\end{tcolorbox} \begin{tcolorbox}[width=0.9\columnwidth, colback=white!95!black]\small $s$ Is it <MASK>? \end{tcolorbox}
    with "subjective" and "objective" as verbalizers.

    \item For \textbf{TREC}, given a question $q$, the task is to classify the type of it:
    \begin{tcolorbox}[width=0.9\columnwidth, colback=white!95!black]\small $q$ <MASK>:  \end{tcolorbox} \begin{tcolorbox}[width=0.9\columnwidth, colback=white!95!black]\small $q$  Q:<MASK>: \end{tcolorbox} \begin{tcolorbox}[width=0.9\columnwidth, colback=white!95!black]\small $q$ why<MASK>?\end{tcolorbox} \begin{tcolorbox}[width=0.9\columnwidth, colback=white!95!black]\small $q$ Answer: <MASK>. \end{tcolorbox}
    with "Description", "Entity", "Expression", "Human", "Location", "Number" as verbalizers for question types of "Description", "Entity", "Abbreviation", "Human", "Location", and "Numeric".

    \item For entailment task (\textbf{RTE}) given a premise $p$ and hypothesis $h$: 
    \begin{tcolorbox}[width=0.9\columnwidth, colback=white!95!black]\small "$h$" ? | <MASK>,  "$p$"\end{tcolorbox} \begin{tcolorbox}[width=0.9\columnwidth, colback=white!95!black]\small $h$? | <MASK>, $p$\end{tcolorbox}\begin{tcolorbox}[width=0.9\columnwidth, colback=white!95!black]\small "$h$" ? | <MASK>. $p$\end{tcolorbox}
    with "Yes" as a verbalizer for entailment, "No" for contradiction.
    
    \begin{tcolorbox}[width=0.9\columnwidth, colback=white!95!black]\small $p$ question: $h$ True or False? answer: <MASK>\end{tcolorbox}
    with "true" as a verbalizer for entailment, "false" for contradiction.

    \item For entailment task (\textbf{CB}) given a premise $p$ and a hypothesis $h$: 
    \begin{tcolorbox}[width=0.9\columnwidth, colback=white!95!black]\small "$h$" ? | <MASK>, "$p$"\end{tcolorbox} \begin{tcolorbox}[width=0.9\columnwidth, colback=white!95!black]\small $h$? | <MASK>, $p$\end{tcolorbox}\begin{tcolorbox}[width=0.9\columnwidth, colback=white!95!black]\small "$h$" ? | <MASK>. $p$\end{tcolorbox}
    with "Yes" as a verbalizer for entailment, "No" for contradiction, "Maybe" for neutral.
    
    \begin{tcolorbox}[width=0.9\columnwidth, colback=white!95!black]\small p question: h true, false or neither? answer: <MASK> \end{tcolorbox}
    with "true" as a verbalizer for entailment, "false" for contradiction, "neither" for neutral.
    
    \item For \textbf{QNLI}, given a sentence $s$ and question $q$:
    \begin{tcolorbox}[width=0.9\columnwidth, colback=white!95!black]\small $s$. Question: $q$? Answer: <MASK>.\end{tcolorbox}
    with "Yes" or "true" as verbalizers for entailment and "No" or "false" for not entailment. 
    \begin{tcolorbox}[width=0.9\columnwidth, colback=white!95!black]\small $s$. Based on the previous sentence, $q$? <MASK>.\end{tcolorbox}
    with "Yes" or "true" as verbalizers for entailment and "No" or "false" for not entailment. 
    \begin{tcolorbox}[width=0.9\columnwidth, colback=white!95!black]\small Based on the following sentence, $q$?<MASK>.$s$\end{tcolorbox}
    with "Yes" and "No" as verbalizers for entailment and not entailment respectively.

    \item For \textbf{QQP}, given two questions $q_1$ and $q_2$:
    \begin{tcolorbox}[width=0.9\columnwidth, colback=white!95!black]\small Do $q_1$ and $q_2$ have the same meaning?<MASK>.\end{tcolorbox}
    with "Yes" or "true" as verbalizers for duplicate and "No" or "false" for not duplicate. 
    \begin{tcolorbox}[width=0.9\columnwidth, colback=white!95!black]\small $q_1$. Based on the previous question, $q_2$? <MASK>. \end{tcolorbox}
    with "Yes" or "true" as verbalizers for duplicate and "No" or "false" for not duplicate. 
    \begin{tcolorbox}[width=0.9\columnwidth, colback=white!95!black]\small Based on the following question, $q_1$?<MASK>.$q_2$ \end{tcolorbox}
    with "Yes" and "No" as verbalizers for duplicate and not duplicate respectively.

    \item For \textbf{MRPC}, given two sentences $s_1$ and $s_2$:
    \begin{tcolorbox}[width=0.9\columnwidth, colback=white!95!black]\small Do $s_1$ and $s_2$ have the same meaning?<MASK>.\end{tcolorbox}
    with "Yes" or "true" as verbalizers for equivalent and "No" or "false" for not equivalent. 
    \begin{tcolorbox}[width=0.9\columnwidth, colback=white!95!black]\small $s_1$. Based on the previous sentence, $s_2$? <MASK>. \end{tcolorbox}
    with "Yes" or "true" as verbalizers for equivalent and "No" or "false" for not equivalent. 
    \begin{tcolorbox}[width=0.9\columnwidth, colback=white!95!black]\small Based on the following sentence, $s_1$?<MASK>.$s_2$ \end{tcolorbox}
    with "Yes" and "No" as verbalizers for equivalent and not equivalent respectively.

    \item For \textbf{WiC}, given two sentences $s_1$ and $s_2$ and a word $w$, the task is to classify whether $w$ is used in the same sense.
    \begin{tcolorbox}[width=0.9\columnwidth, colback=white!95!black]\small "$s_1$" / "$s_2$". Similar sense of "$w$"? <MASK>.\end{tcolorbox}
    \begin{tcolorbox}[width=0.9\columnwidth, colback=white!95!black]\small $s_1$ $s_2$ Does $w$ have the same meaning in both sentences? <MASK> \end{tcolorbox}
    With "No" and "Yes" as verbalizers for False, and True.
    \begin{tcolorbox}[width=0.9\columnwidth, colback=white!95!black]\small $w$ . Sense (1) (a) "$s_1$"  (<MASK>) "$s_2$"  \end{tcolorbox}
    With "2" and "b" as verbalizers for False, and True.
\end{itemize}

\section{Impact of the Position of Masks in Sentence-pair Datasets} \label{app:mask_position}
We evaluate the impact of the position of mask tokens in sentence-pair benchmarks. Given two sentences $s_1$ and $s_2$, we consider the following four locations for inserting mask tokens, where in the case of encoding as two sentences, input parts to the encoder are separated with $|$:
\begin{enumerate}
    \item \drawbox{ $s_1$ $s_2$ <MASK>} \vspace{-0.7em}
    \item \drawbox{ $s_1$ <MASK> $s_2$}\vspace{-0.7em}
    \item \drawbox{ $s_1$ | <MASK> $s_2$}\vspace{-0.7em}
    \item \drawbox{ $s_1$ | $s_2$<MASK>}
\end{enumerate}

Table \ref{tab:mask_positions} shows how the position of masks impact the results. As demonstrated, pattern 2, inserting mask tokens between the two sentences and encoding both as a single sentence obtains the highest validation performance. We use this choice in all the experiments when removing handcrafted patterns.

\begin{table}[tp]
\centering
\begin{adjustbox}{max width=1.0\linewidth}
\begin{tabular}{lllll}
\toprule 
\textbf{Datasets} & \textbf{1} & \textbf{2} & \textbf{3} & \textbf{4}\\
\toprule 
CB &  89.8 & \textbf{91.6}&88.9  &86.5\\
RTE & \textbf{69.1}&\textbf{69.1}&64.5&65.3\\ 
QNLI& 72.0 & \textbf{83.3}&  77.7 &73.1\\
MRPC &  71.6 &69.5 & 66.4 & \textbf{72.0}\\ 
QQP &  79.2&\textbf{82.8}&72.5&70.2 \\
WiC & \textbf{60.3}&59.5&60.2& 59.5\\
\midrule 
Avg &73.7 &\textbf{76.0}  &71.7& 71.1 \\
\bottomrule
\end{tabular}
\end{adjustbox}
\caption{Validation performance for sentence-pair benchmarks for different locations of mask tokens. Bold fonts indicate the best results in each row.} 
\label{tab:mask_positions}
\end{table}

\section{Impact of Initialization}\label{app:sigma}
We initialize the label embedding matrix with random initialization from a normal distribution $\mathcal{N}(0, \sigma)$. In table \ref{tab:noise}, we show the development results for different values of $\sigma$. We choose the $\sigma$ obtaining the highest performance on average over average and worst case performance, i.e., $\sigma=10^{-4}$.

\begin{table}[tp]
\centering
\begin{adjustbox}{max width=1.0\linewidth}
\begin{tabular}{l@{\hskip 0.1in}l@{\hskip 0.1in}l@{\hskip 0.1in}l@{\hskip 0.1in}l}
\toprule 
\textbf{Datasets} & \textbf{$10^{-2}$} & \textbf{$10^{-3}$} & \textbf{$10^{-4}$} & \textbf{$10^{-5}$}\\
\toprule 
CB    &90.0\smaller{/82.5} &92.2\smaller{/85.0}&91.6\smaller{/87.5} &91.6\smaller{/87.5} \\
MRPC  &69.8\smaller{/56.2}&70.8\smaller{/56.2}&69.5\smaller{/56.2}&70.8\smaller{/56.2}\\
QNLI  &83.3\smaller{/71.9}&82.7\smaller{/71.9}&83.3\smaller{/71.9}&83.1\smaller{/68.8}\\
QQP   &82.8\smaller{/78.1}&82.7\smaller{/75.0}&82.8\smaller{/75.0}&83.0\smaller{/75.0}\\
RTE   &69.8\smaller{/62.5}&69.2\smaller{/59.4}&69.1\smaller{/62.5}&68.3\smaller{/62.5}\\ 
WiC   &62.2\smaller{/50.0}&59.7\smaller{/46.9}&59.5\smaller{/53.1}&58.9\smaller{/50.0}\\
\midrule 
Avg   &76.3\smaller{/66.9}&76.2\smaller{/65.7}&76.0\smaller{/67.7}&76.0\smaller{/66.7} \\
\midrule 
Total Avg & 71.6&  71.0   &  \textbf{71.8}                 & 71.3 \\
\bottomrule
\end{tabular}
\end{adjustbox}
\caption{Validation performance for different values of $\sigma$. We show mean performance/worst-case performance across 20 runs. The last row shows the average of mean performance/worst-case performance.}
\label{tab:noise}
\end{table}

\begin{table*}
\centering 
\begin{tabular}{l|@{\hskip 0.05in}l@{\hskip 0.15in}l@{\hskip 0.15in}lHl}
\toprule 
\textbf{Dataset} & \textbf{\perfect} & \textbf{-Hinge Loss} & \textbf{+Label Emb} & \textbf{-Adapters+BitFit}& \textbf{-Prototypical} \\
\midrule 
SST-2 & \textbf{90.7\small{/88.2}}\small{/1.2} & 90.0\small{/85.9/1.7} & 90.6\small{/87.6/\textbf{1.1}}&89.5\small{/81.7/3.0}&90.4\small{/85.2/1.6}\\
CR    & 90.0\small{/85.5/1.4} &\textbf{90.1\small{/88.6/0.9}} &89.7\small{/86.6/1.4}&90.1\small{/87.8/1.0}&89.9\small{/86.8/1.4}\\
MR    & \textbf{86.3}\small{/81.4/1.6} & 85.2\small{/78.6/2.4} &85.8\small{\textbf{/82.4/1.4}}&85.6\small{/80.5/1.9}&85.7\small{/78.0/2.0}\\
SST-5 & 42.7\small{/35.1/2.9} &\textbf{43.3}\small{/36.8/3.1} &41.8\small{/\textbf{37.1}/2.5}&42.3\small{/36.8/3.3}&41.2\small{/35.9/\textbf{2.4}}\\
SUBJ  & 89.1\small{/82.8/2.1} &89.4\small{/83.1/2.2} &\textbf{90.0\small{/86.0/1.8}}&89.1\small{/82.4/2.4}&89.7\textbf{\small{/86.0/1.8}}\\
TREC  & \textbf{90.6\small{/81.6/3.2}} &89.9\small{/76.8/4.2} &89.7\small{/71.6/6.1}&90.4\small{/85.0/1.4}&89.6\small{/76.2/4.9}\\
CB    &\textbf{90.3\small{/83.9}/3.5} &89.2\small{/80.4/4.8} &89.6\small{/82.1/3.6}&89.6\small{/82.1/4.3}&89.3\small{/80.4/3.9}\\
RTE   & 60.4\small{/53.1/4.7}&\textbf{60.7\small{/54.5/4.0}}
&58.6\small{/50.9/4.0}&61.3\small{/53.8/5.2}&58.5\small{/50.9/4.5}\\ 
QNLI  & 74.1\small{/60.3/4.6}&72.9\small{/64.4/3.9} &\textbf{74.9}\small{/66.7/3.6}&70.6\small{/51.9/5.9}&74.7\small{/\textbf{67.5/3.5}}\\ 
MRPC  & 67.8\small{/54.7/5.7}&67.0\small{/49.8/5.5}&\textbf{68.1\small{/56.9/4.8}}&68.5\small{/57.4/5.1}&\textbf{68.1\small{/56.9/4.8}}\\ 
QQP   &\textbf{71.2\small{/64.2/3.5}} &69.9\small{/63.0/4.1}&70.3\small{/62.2/4.0}&69.4\small{/63.0/3.9}&70.2\small{/62.2/4.0}\\ 
WiC   &\textbf{53.8}\small{/47.0/3.0} &53.7\small{/46.7/3.3}&53.6\small{/\textbf{50.2/2.4}}&52.9\small{/47.8/2.7}&53.6\small{/50.0/2.6}\\
\midrule 
Avg & \textbf{75.6/}\small{68.1/\textbf{3.1}} & 75.1\small{/67.4/3.3} &75.2\small{/\textbf{68.4/3.1}}&74.9\small{/67.5/3.3}&75.1\small{/68.0/\textbf{3.1}}\\ 
\bottomrule
\end{tabular}
\caption{Ablation results on the impact of different design choices in \emph{\perfect}. We report the average performance/worst-case performance/and the standard deviation.}
\label{tab:perfect_ablations} 
\end{table*}

\section{Ablation Results} \label{app:ablation}
To study the impact of different design choices in \perfect, we considered the following experiments:
\begin{itemize}
    \item \textbf{-Hinge Loss:} In this variant, we replace the hinge loss with multi-class cross entropy loss. 
    \item \textbf{+Label Emb:} We use the trained label embeddings during the inference, substituting the computed prototypes in \eqref{eqn:centroids}.
    \item \textbf{-Prototypical:} Instead of using prototypical networks, during inference, we use the same objective as training, i.e.,~\eqref{eqn:total_loss}. 
\end{itemize}

Results are shown in Table~\ref{tab:perfect_ablations}. Experimental results demonstrate that \perfect obtains the best results on average. Using multi-class cross-entropy instead of hinge loss, obtains substantially lower minimum performance (67.4 versus 68.1), demonstrating that training with hinge loss makes the model more stable. Using the trained label embeddings (+Label Emb) obtains very close results to \perfect (slightly worse on average and slightly better on the minimum performance). Using the similar objective as training with replacing prototypical networks (-Prototypical), obtains lower performance on average (75.1 versus 75.6). These results confirm the design choices for \perfect.  
\end{document}